\documentclass[letterpaper]{article}
\usepackage{aaai}
\usepackage{times}
\usepackage{helvet}
\usepackage{courier}
\usepackage{graphicx}
\usepackage{times}
\usepackage{latexsym}
\usepackage{booktabs}
\usepackage{amsmath}
\usepackage{amssymb}
\usepackage{graphicx}
\usepackage{bm}
\usepackage{CJKutf8}
\usepackage{amsmath}
\usepackage{url}
\usepackage{tabularx}
\usepackage{subfig}
\usepackage{multirow}

\usepackage{color}
\usepackage[ruled,linesnumbered]{algorithm2e}
\begin{document}
%
\title{MetaAID 2.0: An Extensible Framework for Developing Metaverse Applications via Human-controllable Pre-trained Models}
\author{
Hongyin Zhu\\
hongyin\_zhu@163.com\\
}
\maketitle
\begin{CJK*}{UTF8}{gbsn}
\begin{abstract}
Pre-trained models (PM) have achieved promising results in content generation. However, the space for human creativity and imagination is endless, and it is still unclear whether the existing models can meet the needs. Model-generated content faces uncontrollable responsibility and potential unethical problems. This paper presents the MetaAID 2.0 framework, dedicated to human-controllable PM information flow. Through the PM information flow, humans can autonomously control their creativity. Through the Universal Resource Identifier extension (URI-extension), the responsibility of the model outputs can be controlled. Our framework includes modules for handling multimodal data and supporting transformation and generation. The URI-extension consists of URI, detailed description, and URI embeddings, and supports fuzzy retrieval of model outputs. Based on this framework, we conduct experiments on PM information flow and URI embeddings, and the results demonstrate the good performance of our system. 
\end{abstract}

\section{Introduction}

In the past two years, artificial intelligence-generated content (AIGC) has received widespread attention. AIGC is seen as the next step for professionally generated content (PGC) and user-generated content (UGC). The text-to-image model can draw beautiful images in the blink of an eye. ChatGPT \cite{openai2022chatgpt} can bring users high-quality chatting and creative writing. On the one hand, the advancement of pre-trained models (PM) has driven the vigorous development of industries such as artificial intelligence, cloud computing, and big data. On the other hand, models may have the potential for misuse, irresponsibility, or unethical issues. Against this background, we introduce MetaAID 2.0, a framework for developing Metaverse applications. MetaAID 2.0 is an updated version of MetaAID 1.0 \cite{zhu2022metaaid}, dedicated to supporting a flexible, ethical, and human-controllable process of generating Metaverse content. MetaAID 2.0 further expands the capabilities of multimodal data generation and provides a mechanism to control human creativity and responsibility.


Existing systems mainly provide AI services based on large-scale cloud computing platforms, such as AI drawing \cite{ramesh2022hierarchical}, AI writing \cite{openai2022chatgpt}, etc. Few lightweight frameworks support multimodal generation with controllable creativity and responsibility. With the growth of the related open-source community, it has become possible to accomplish this task. The concept of Model-as-a-Service (MaaS) enables end-to-end processing of multimodal data. The idea of the MetaAID 1.0 framework is to form a good collaborative relationship between AI technology and human editors. MetaAID 2.0 continues this philosophy, adding the concept of human-controllable AI given the rapid growth in the capabilities of pre-trained models. Human controllability can be divided into two dimensions: the controllability of human creativity and the controllability of model outputs responsibility. Through the PM information flow \cite{zhu2022switchnet} set by users, the controllability of human creativity can be realized, the value of human imagination can be maximized, and cultural innovation can be carried out. By generating URI-extension for output resources, responsibility for output data can be controlled, enabling ``data ownership" by the way. Our framework supports receiving multimodal input (text, image, video, and audio), unleashing human creativity through PM information flows and enabling responsible and ethical interaction of model outputs through URI-extension.

Through the customized PM information flow, it is possible to control human creativity, create rich and colorful metaverse content, and realize the input and output of multimodal data. The challenge of this problem is that the framework needs to contain rich functions and be able to handle multimodal data flexibly. For example, good articles can be written through text generation models, but it faces the limitation of not being vivid. Adding pictures to the article will enhance the ornamental and artistic value of the work. Additionally, adding voice can improve the user experience. Furthermore, showing in the form of generated videos can gain attention on video platforms and generate real-world feedback. We hope to unleash the value of human creativity through PMs and propose PM information flows.


The second challenge is the controllable responsibility of the model. Although image and video digital watermarking techniques \cite{mohanarathinam2020digital} have been extensively studied in prior works, the generated data will inevitably be modified and edited, and even digital watermarks may be removed. This poses a challenge to the controllable responsibility of model data. In addition, the nature of text data is inconvenient to add digital watermarks, and the modified content is more difficult to identify. We hope to resolve this problem through multimodal representation and retrieval. Traditional methods are not easy to achieve fast retrieval and identification of modified data. Inspired by the Semantic Web, we introduce a semantic representation-based knowledge base for controllable responsibility. We propose a URI-extension consisting of URI, detailed description, and URI Embeddings for multimodal data, to realize retrieval and identification.

To evaluate this framework, we constructed an experimental website. The innovations of this paper are as follows:

(1) We introduce MetaAID 2.0, which aims to realize the controllability of human creativity and the controllability of model outputs.

(2) We propose PM information flow and URI-extension mechanisms. On the input side, users can give full play to their creativity by controlling the PM information flow. On the output side, the framework realizes controllable data responsibility and ethical interaction through URI-Extension.


(3) We develop a website based on this framework and achieve good application results.

\section{Related Work}
\subsection{Single-Modal Pre-trained Models}
ChatGPT \cite{openai2022chatgpt,ouyang2022training} is a chatbot program developed by OpenAI, which uses a language model based on the GPT-3.5 architecture and is trained by reinforcement learning. They collected a dataset of model output rankings and further fine-tuned the model through reinforcement learning with human feedback.
Google proposes Bard which is built based on the large language model LaMDA \cite{thoppilan2022lamda}. LaMDA is trained on large-scale online data to generate convincing responses to user prompts, and the model can remove unsafe answers and select high-quality responses.

Schick et al. present Toolformer\cite{schick2023toolformer}, which learns to call different tools such as search engines, calculators, and translation systems using APIs in a self-supervised manner. Toolformer can select the APIs, timing, and parameters to be called after training. 
OpenAI's Brown et al. \cite{brown2020language} pre-train GPT-3 which contains 175 billion parameters. GPT-3 requires no fine-tuning and achieves good performance on zero-shot, one-shot, and few-shot NLP tasks. 
Baidu's Wang et al. \cite{wang2021ernie} build ERNIE 3.0 Titan which contains 260 billion parameters. Based on the ERNIE 3.0 framework, this model achieved state-of-the-art results on 68 datasets. With a self-supervised adversarial loss and a controllable language modeling loss, the model is improved in generating believable and controllable text.

\subsection{Multi-Modal Pre-trained Models}
%
Rombach et al. \cite{rombach2021highresolution} propose a latent diffusion model to carry out the diffusion process in the latent space, thus greatly reducing the computational complexity. The latent diffusion mechanism can significantly improve the training and sampling efficiency, and the model uses a cross-attention mechanism for a conditional generation. 
DALL-E 2 \cite{ramesh2022hierarchical} combines a CLIP-based text encoder with an image embedding decoder to construct a complete text-to-image model. The proposed two-stage model first generates CLIP image embeddings from text descriptions and then uses a decoder to generate images based on the image embeddings. 

Agostinelli et al. propose MusicLM \cite{agostinelli2023musiclm}, a model for generating music from text or other signals, which can quickly generate high-quality music at 24 kHz according to conditional input. The model treats the conditional music generation process as a hierarchical sequence-to-sequence modeling task.
Shah et al. \cite{shah2022lm} propose large model navigation LM-Nav, which realizes mobile robot navigation based on text instructions. LM-Nav combines three pre-trained models for parsing user commands, estimating the probability of observations matching those landmarks, and estimating navigational features and robot actions. 


\section{Pre-trained Models Information Flow}

\subsection{Functional module architecture}
\begin{figure*}[!h]
\centering
\includegraphics[width=4in]{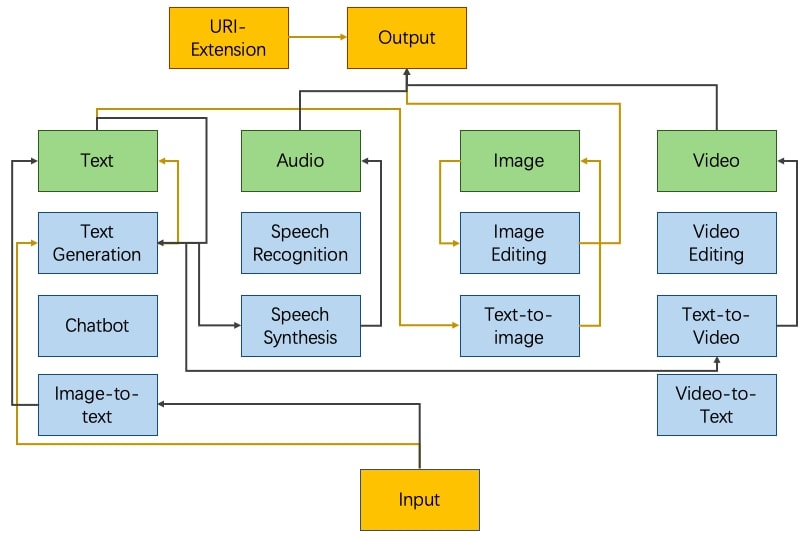}
\caption{Schematic diagram of Our PM information flows}
\label{arch.fig}
\end{figure*}

In this subsection, we will introduce the PM modules in the framework, each named after some Beijing locations where I have lived. To realize the controllability of human creativity, we design different PM information flows \cite{zhu2022switchnet}. As shown in Figure \ref{arch.fig}, the modules in it can be combined in a custom way. Let's demonstrate with 2 PM information flows as an example.

(1) The initial text is extended by a text generation module, and then the content is expressed as a description suitable for image generation by manually adjusting the generated text. Text-to-image models can generate images based on conditional text input. Further, in order to add an artistic effect to the image, the original image is fused with world-famous paintings such as Van Gogh's ``The Starry Night" to achieve style transfer, and finally, the image output is obtained, and a URI-extension is generated at the same time.

(2) Image-to-text models convert input images into textual descriptions. Then, the text is fed into a text generation model for expanded creative content. Manually polish the text content and let the speech synthesis model generate audio files. At the same time, part of the text is also used to generate video files. Finally, the system outputs audio and video files and generates the corresponding URI-extension.

As a tool to assist humans in developing applications, its flexibility and rich functions are crucial. The framework strives to unleash human creativity and break functional constraints. Specific data modalities use corresponding models, and this framework realizes the transformation and generation between various modes through PM information flows.


\subsection{Text Modal}
We introduce the Zhongguancun (中关村) text module. This module mainly consists of text generation and Chatbot.
\subsubsection{Text generation}
Text generation aims to generate fluent natural language text based on input data \cite{li2021pretrained}. 
We introduce Chinese-English bilingual text generation models. For English, OpenAI's Radford et al. \cite{radford2019language} propose GPT-2 which is a text generation model based on a transformer decoder. The transformer decoder is mainly composed of three components, multi-head self-attention mechanism, masked multi-head self-attention mechanism, and fully connected feedforward network. GPT-2 is an autoregressive model, and the model generates a word every iteration during the computing process. A language model can be expressed as Equation (1)
\begin{align}
p(x) = \prod_{i=1}^n p(x_i|x_{i-1},...,x_1)
\end{align}
where x is the input text and $x_i$ represents the i-th word. When GPT-2 performs zero-shot learning, it also regards the task form as an input, and can directly predict the answer of the downstream task without fine-tuning the model. Similarly, few-shot learning takes the form of using the task template and a small number of examples as input to guide the model to directly predict the answers to downstream tasks.

GPT-2 first pre-trains a language model on large-scale text data and then applies the language model directly to a variety of supervised downstream tasks without the need for task-specific fine-tuning. By designing specific templates for various tasks, the generated texts are used as prediction results. This form of zero-shot learning gradually developed into the paradigm of ``prompt learning" \cite{liu2023pre} and was formally proposed. 

For Chinese, Zhao et al. \cite{zhao2019uer} propose a UER toolkit and train Chinese models. The toolkit supports pre-training and fine-tuning and has loosely coupled and encapsulated rich modules.

\subsubsection{Chatbot}
Chatbots can converse with users through natural language and understand their intent. A text generation model can support a dialogue system. We have implemented a Pre-trained language model (PLM)-based dialogue system which supports bilingual chat in Chinese and English.

For English, Roller et al. \cite{roller2020recipes} propose an open-domain chatbot, BlenderBot. They let three different models jointly decide whether to introduce knowledge by combining a retriever and a generative model. Finally, better results are achieved through the Beam Search decoding strategy. 

For Chinese, we use ChatYuan\footnote{https://github.com/clue-ai/ChatYuan}, which is based on PromptCLUE\footnote{https://github.com/clue-ai/PromptCLUE}. For the training process, ChatYuan used a combined dataset of large-scale functional dialogues and multi-round dialogues. PromptCLUE is built on the T5 model, which is pre-trained for large-scale Chinese corpus and performs multi-task learning on more than 100 tasks. The Chinese language understanding evaluation (CLUE) \cite{xu2020clue} benchmark is established in 2019 and corresponds to the foreign General Language Understanding Evaluation (GLUE) \cite{wang2018glue} benchmark.

\subsection{Audio Modal}
We present the Houhai (后海) audio module which consists of speech recognition and speech synthesis functions.
\subsubsection{Speech Recognition}
The audio technology module mainly includes the functions of speech-to-text and text-to-speech. 
For Chinese and English, Li et al. \cite{li2020robutrans} improved transformerTTS \cite{li2019neural} and proposed RobuTrans. The data features of the encoder include phoneme and prosodic features, which make the synthesized audio more natural and use duration-based hard attention to effectively improve the robustness of the model. Some companies provide high-quality speech recognition APIs, such as Baidu, HKUST Xunfei, etc.

\subsubsection{Speech Synthesis}
Due to the significant difference in the pronunciation synthesis of different languages, we use different models for different languages. For Chinese, Gao et al. \cite{gao2022paraformer} proposed a single-round non-autoregressive model Paraformer. The predictor module is used to predict the number of target characters in the speech, and the sampler transforms the acoustic feature vector and the target character vector into a feature vector containing semantic information.
For English, Gao et al. \cite{gao2020universal} propose universal ASR which supports both streaming and non-streaming ASR models. 

\subsection{Image Modal}
We present the Wudaokou (五道口) image module composed of image editing and image generation.
\subsubsection{Image Editing}
Image editing refers to the processing and modification of the original image so that the image meets the needs of a specific scene. Image enhancement focuses on transforming digital images into a state more suitable for display or analysis. 

GAN can generate excellent texture information, but it is unstable and will generate false textures. Liang et al. \cite{liang2022LDL} propose an image super-resolution model which explicitly distinguishes GAN-generated pseudo-textures from real details. They propose a local discriminative learning (LDL) framework to regularize the adversarial training process. 

Deng et al. \cite{deng2021stytr2} propose a style transfer transformer $StyTr^2$ which uses two transformer encoders to encode content and style features. Different from the auto-regressive processing method, they use a decoder to predict the output using all consecutive image patches as input.
Kulshreshtha et al. \cite{kulshreshtha2022feature} propose a multi-scale refinement technique which is an iterative refinement method from coarse to fine to improve the inpainting quality of neural networks at high resolution.
Image modification with stable diffusion \cite{rombach2021highresolution} achieves promising results. This model is implemented based on the latent diffusion technique, which can reduce memory and computational complexity. The model mainly consists of three parts, autoencoder, U-Net, and text encoder, and is trained to generate a latent (compressed) representation of the image. 

\subsubsection{Image Generation}
Text-to-image models take a textual description as input and output a corresponding image that matches the input. This module contains different open-source image generation models, such as DALL·E, DALL·E 2, Stable diffusion, Imagen, etc. to generate images in various styles.

OpenAI's Ramesh et al.\cite{ramesh2022hierarchical} propose DALL·E 2 which realizes the text-conditioned image generation process through a two-stage model. First, the input text is encoded through the CLIP model and converted into a representation of the image features, and then the final image is generated by a decoder according to the image features. The image decoder is learned to reverse the image encoding process via the GLIDE model (diffusion model) to randomly decode the CLIP image embedding. Google's Saharia et al. propose Imagen \cite{saharia2022photorealistic} which has a simple and powerful process of generating images from text. First, the diffusion model is used to generate images from text, and then the model uses two super-resolution diffusion models based on the obtained images to achieve image resolution enhancement.

Since these models only support English input, to meet the needs of multilingual image generation, We integrate machine translation and prompt expansion functions to achieve multilingual image generation tasks. The prompt expansion process is based on the semantic retrieval of the prompt keyword knowledge base. Images generated without expanded descriptions cannot achieve satisfactory results. It is not easy for ordinary users to master the skills of using keywords, but with this function, users' efficiency is greatly improved.

\subsubsection{Image-to-Text}
Google's Dosovitskiy et al. \cite{dosovitskiy2020image} propose a vision transformer (ViT) that applies transformer to image classification tasks. GPT-2 is a language model that can be used to generate text. These two models can be combined to form the visual encoding and decoding process. The combination of the two models can be achieved via the cross-attention mechanism, which enables the decoder to retrieve key information from the encoder. 

\subsection{Video Modal}
We introduce the Gulou (鼓楼) video module which consists of video editing and video generation functions.
\subsubsection{Video Editing}
The video technology module is designed to edit, transform, enhance, and generate video files. 
Petrov et al. \cite{perov2020deepfacelab} propose to use deepfake \cite{nguyen2022deep} technology for face swapping, realize model learning through face detection, face alignment, and face segmentation processes, and provide a set of adjustments tools for human-computer interaction.
Chan et al. \cite{chan2022investigating} propose a video super-resolution model, which consists of a stochastic degradation process that uses longer sequences rather than larger batches during training, which allows for more efficient use of temporal information and more stable performance. Besides, an image pre-cleaning module is proposed to balance detail synthesis and artifact suppression. 

Narasimhan et al. propose CLIP-It \cite{narasimhan2021clip} which generates a video summary from either a user-defined natural language description or an automatically generated dense video description based on the original video. The language-guided attention head fuses image and language embeddings, and the transformer for video frame scoring attends to all frames to compute their relevance scores. During inference, video summaries are constructed by converting video frame scores to shot scores and using a knapsack algorithm to select shots with high scores.

\subsubsection{Video Generation}
Text-to-video aims to generate video directly based on the natural language description. 
Imagen video \cite{ho2022imagen} is a text-conditioned video generation system based on the cascaded video diffusion model, which uses a basic video generation model and spatial-temporal super-resolution models to generate high-definition videos. They propose to extend the text-to-image Imagen model \cite{saharia2022photorealistic} in the temporal domain to migrate to video generation.
Make-A-Video \cite{singer2022make} extended the diffusion-based text-to-image model to text-to-video by spatiotemporally decomposing the diffusion model. They spatially and temporally approximate the decomposed temporal U-Net and attention tensor and generate high-definition, high-frame-rate video through super-resolution methods. Hong et al. propose CogVideo \cite{hong2022cogvideo} which contains a hierarchical learning method with multiple frame rates. CogVideo is built on top of the CogView2 model \cite{ding2022cogview2} used for image generation. They propose a spatiotemporal bi-directional attention mechanism to learn smooth temporal video frame information.

\section{Human-Controllable Mechanisms}
\subsection{Universal Resource Identifier Extension}
In the Semantic Web, a Universal Resource Identifier (URI) is defined as a unique string used to identify a resource, which can represent any object or abstract concept. We introduce URI-extension consisting of 3 elements, a URI, a detailed description, and URI embeddings. In terms of working mechanisms, this technology leverages semantic representation, high-performance storage, and semantic retrieval. We will describe the 3 parts of the URI-extension.

(1) Inspired by the URI in the Semantic Web, we also take the form of URIs, aiming at generating URIs for each resource to uniquely locate the outputs. 

(2) Detailed description is used to add more descriptions to facilitate tracking of the framework output. This problem arises in the context of big data. Many users use AI to generate outputs in different locations and at different times. The detailed description contains more information that is convenient for positioning and analysis, and the content of this description is relatively long.

(3) URI embeddings are used for efficient identification and retrieval. This technology is used for the convenience of information comparison, and identification to realize efficient retrieval. It is inevitable for people to modify the framework output. Even with slight modifications, it is difficult to retrieve from a large amount of output using traditional search methods. We hope to extract the implicit semantic representation through pre-trained models, to deal with fuzzy retrieval.


The working mechanism consists of two parts, URI-extension, and semantic retrieval. Detailed descriptions and semantic representations are stored in a lightweight fashion.
The resource information extraction algorithm is designed to generate unique URIs for the model output. we add information such as device, IP address, user account, date, etc. to generate detailed descriptions. Next, we will further describe the generation of multimodal URI embeddings. 

\subsection{Multimodal URI Embeddings}

This part introduces the method to generate part (3) of URI-extension, by which URI-extension can be retrieved efficiently. 
For images, we adopt the ViT model \cite{wu2020visual} to generate vector representations. ViT employs a convolutional neural network (CNN) to convert an image into a sequence of image patches, thereby utilizing the transformer model to produce a vector representation. This vector representation contains the structural information and content information of the image.

An acoustic fingerprint is a compressed representation of an audio signal that can be used to identify audio samples or retrieve similar audio files in a database. Audio fingerprinting 
\cite{borkar2021music} aims to generate vector representations for audio. We employ this technique to generate audio representations.

Huang et al. \cite{huang2021self,qing2022hico} propose a method to learn high-quality video representation. This method separates static context and motion prediction, including the task of learning coarse-grained representations for context matching, and the task of learning fine-grained representations for motion prediction.

Sentence-BERT \cite{reimers-2019-sentence-bert} feeds a text sequence into a BERT model to obtain an overall representation vector, which is used for downstream tasks such as retrieval and clustering.

URI, detailed descriptions, and URI-embeddings are kept corresponding. Their efficient management is a challenging problem. Considering that management will become difficult as the amount of data increases, we do not use a general database. For vector representation, it is not suitable to use a database for storage and computation. We use flexible serialization tools for the persistent storage of data. 
Parallel acceleration can be achieved through tensor computation which facilitates fast semantic retrieval.

\section{Experiment}

\subsection{Setup}
We developed a website based on this framework and invited users to test functions. This application runs on an AMD Ryzen-9 5900X Processor @ 3.7GHz (Mem: 128G) and RTX 4080 GPUs (16G).

\subsection{Results of PM Information Flow}
This section takes the information flow shown in Figure \ref{arch.fig} as an example to illustrate the processes including text generation, image generation, audio generation, and video generation. Let's take the story of ``AI and humans falling in love" as an example to illustrate the ability of PM.

(1) To generate a script, we use a GPT-2 to write movie scripts. Then based on this movie script, the movie plot is optimized manually.

\texttt{
The artificial intelligence becomes a beautiful woman Meggie calling for help, and the man Jack enters the virtual world. Jack was killed in the virtual world. A few months later, another woman joined Meggie, her name is ``Aries". However, Jack is rescued by a woman named ``Cynthia", who is the companion to The Immortal Mother...}

(2) To generate audio, we use the speech synthesis module for the narration of the story. If the movie script contains dialogues of characters, we can also use the speech synthesis model to generate dialogues of characters with different timbres.

\begin{figure*}[!htbp]
\centering
\includegraphics[width=6.5in]{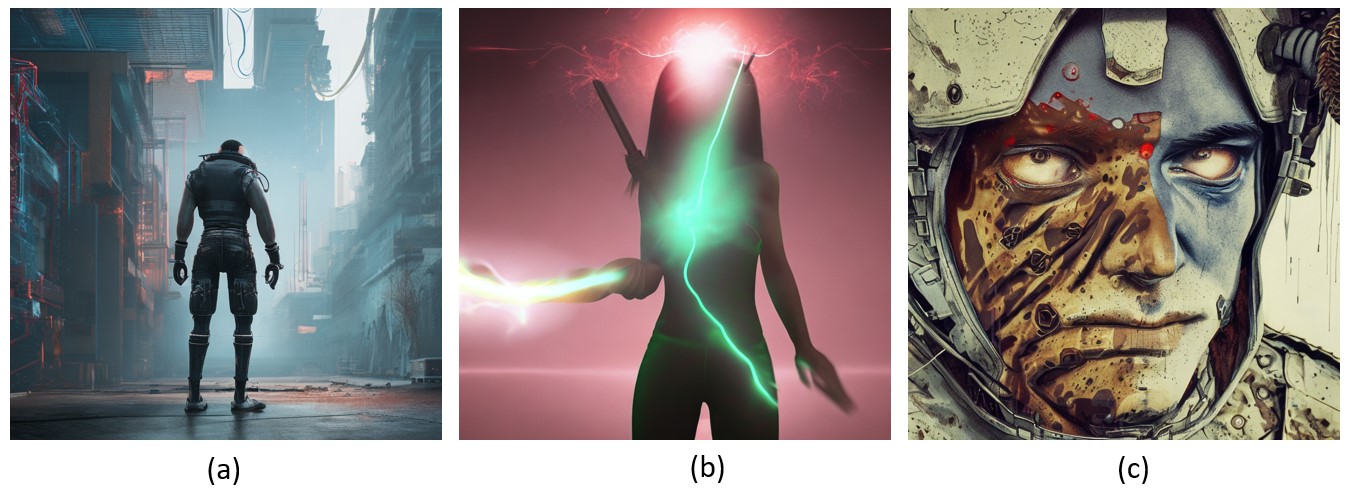}
\caption{Scenes generated for the movie plot}
\label{person.fig}
\end{figure*}

(3) We use a text-image model to generate scenes for the movie, including natural environments, backgrounds, characters, tools, etc., as shown in Figure \ref{person.fig}.

\begin{figure*}[!htbp]
\centering
\includegraphics[width=6.5in]{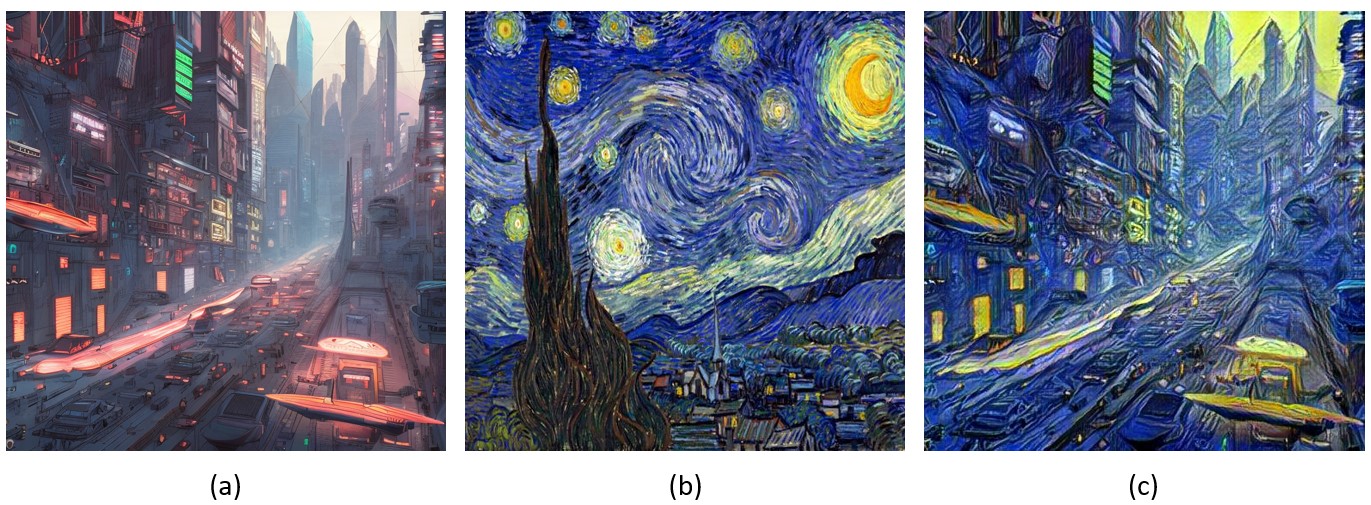}
\caption{Adding Van Gogh's ``The Starry Night" style to a virtual city with style transfer}
\label{cyber.fig}
\end{figure*}

(4) Sometimes it is necessary to add an image style to make this video look more interesting, as shown in Figure \ref{cyber.fig}.

\begin{figure*}[!htbp]
\centering
\includegraphics[width=6.5in]{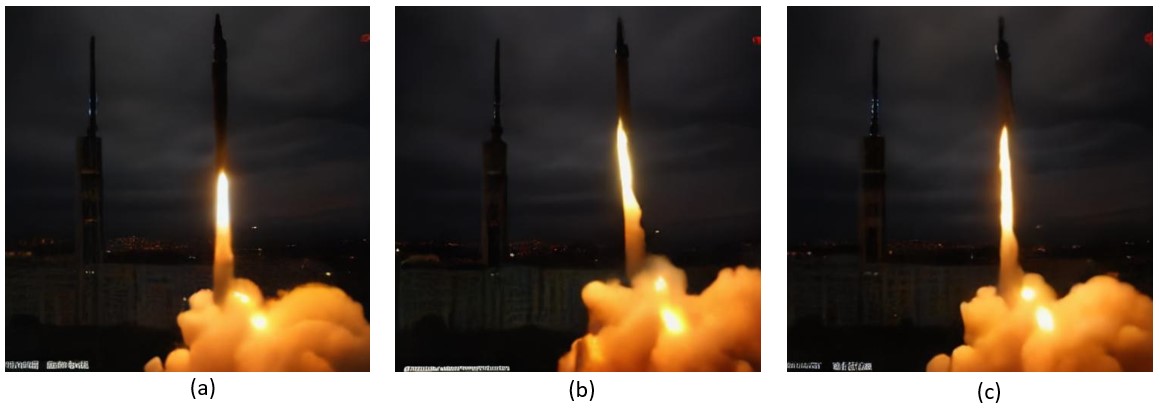}
\caption{Dynamic content generated for the movie plot}
\label{video.fig}
\end{figure*}

(5) For shots with dynamic content, e.g., the moment a person falls, the process of skiing, etc., the text-to-video module is used for the generation, as shown in Figure \ref{video.fig}.

(6) Finally, we compose a video based on a series of materials. We can combine the above materials through programmatic operations, or for convenience, we use video editing software to combine them.

\subsection{Results of URI Embeddings}
\begin{figure}[!htbp]
\centering
\includegraphics[width=3in]{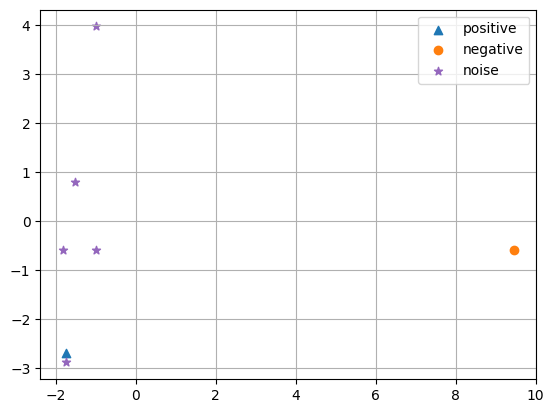}
\caption{A reduced dimensional representation after adding noise to the image, where the positive sample represents the original image, the negative sample represents a different image used for comparison, and the noise samples represent the modified image}
\label{pca.fig}
\end{figure}

This section mainly shows the experimental results of URI embeddings. The similarity is computed based on the embedding representations. We use image and text data as demonstrations. For image data, we have made significant changes to the original image, such as masking regions in the image and adding new content. Then we use PCA to visualize the URI embeddings, as shown in Figure \ref{pca.fig}. We can see that the URI embedding maintains the similarity of the same image.

\begin{figure}[!htbp]
\centering
\includegraphics[width=3in]{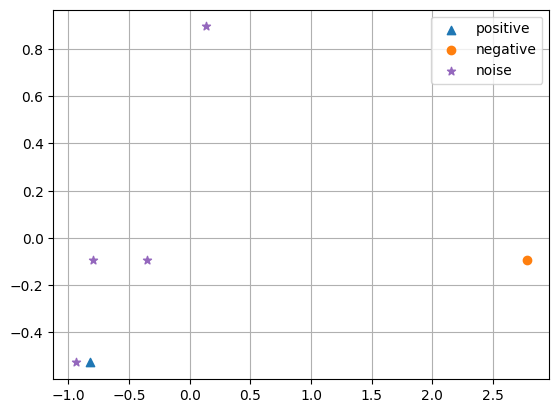}
\caption{A reduced-dimensional representation with noise added to the movie plot, where the positives represent the original movie plot, the negatives represent another textual description used for comparison, and the noise samples represent the modified movie plots}
\label{text.fig}
\end{figure}

We made obvious changes to the text data, such as removing several parts, adding other content, switching the order, changing sentences, etc. Then we visualized the URI embeddings with PCA as shown in Figure \ref{text.fig}. We can see that the URI embedding maintains the similarity of the same text.

\section{Conclusion and future work}
This paper introduces MetaAID 2.0, which implements human-controllable PM information flow. The controllability of human creativity is achieved through multimodal PM information flow. The controllability of model responsibilities is achieved through URI-extension. The multimodal PM information flow in this paper covers processing modules for text, image, video, and audio data.

This paper mainly introduces the human-controllable PM information flow. In the future, we hope to integrate this framework with real-world application scenarios to empower the digital economy.

\end{CJK*}

\clearpage
\bibliographystyle{aaai}
\bibliography{reference}

\end{document}